\documentclass[letterpaper, 10 pt, conference]{ieeeconf}  

\IEEEoverridecommandlockouts                              

\usepackage{amsmath} 
\usepackage{amssymb}  
\usepackage{amsfonts}

\usepackage{multicol}
\usepackage{bm}
\usepackage{graphicx}
\graphicspath{{img/}}
\usepackage{xcolor}
\usepackage{mathtools}
\usepackage{mathabx}
\usepackage{siunitx}
\usepackage{subcaption}
\usepackage{graphicx}  
\usepackage{pgfplots}
\pgfplotsset{
compat=newest,
every tick label/.append style={scale=0.8},
every axis label/.append style={scale=0.8}
}

\usetikzlibrary{plotmarks}
\usetikzlibrary{arrows.meta}
\usepgfplotslibrary{patchplots}
\usepackage{grffile}

\usepackage{dashbox}
\newcommand\dashedph[1][H]{\setlength{\fboxsep}{0pt}\setlength{\dashlength}{2.2pt}\setlength{\dashdash}{1.1pt} \dbox{\phantom{#1}}}

\usepackage{hyperref}
\hypersetup{
    pdfborder={0 0 0},
}

\usepackage{xcolor}

\newcommand{\ssd}{\textrm{d}} 
\newcommand{\sstd}{\textrm{td}} 
\newcommand{\smax}{\textrm{max}} 

\title{\LARGE \bf
Gray-Box Model Predictive Control for Articulated Dump Trucks via Gaussian Process Learning of Sideslip
}

\author{Arash Shahirpour$^{1}$, Jens Ahlers$^{1}$, Christopher Schulte$^{1}$, and Tim Reuscher$^{1}$
\thanks{$^{1}$Arash Shahirpour, Jens Ahlers, Christopher Schulte, and Tim Reuscher are with the Institute of Automatic Control, RWTH Aachen University, 52074 Aachen, Germany\newline{\tt\small { 
\ \{a.shahirpour,j.ahlers, c.schulte, t.reuscher\}}@irt.rwth-aachen.de}}}

\usepackage[nameinlink]{cleveref}

\begin{document}

\maketitle
\thispagestyle{empty}
\pagestyle{empty}

\begin{abstract}
The growing demand for automation in the mining industry, particularly for the autonomous operation of articulated dump trucks (ADTs), has drawn increased attention to accurate vehicle modeling. The importance of such models lies in their use in model predictive control (MPC), model-based estimation methods, and vehicle simulation. While dynamic modeling offers a viable solution for these purposes, it is associated with complex setup and parametrization and may require recalibration in changing operating environments. As a result, kinematic models have dominated ADT modeling, especially in MPCs, at the expense of reduced prediction accuracy.

In this work, we propose an approach using Gaussian Process Regression (GPR) to learn the sideslip angle of the vehicle, which is identified as the primary contributor to the reduced accuracy of kinematic models. The learned GPR function is augmented into the kinematic model to form a gray-box model that aims to reduce the gap to dynamic models. We show that the gray-box model can predict the sideslip angle and, consequently, the vehicle's lateral velocity, thereby improving the MPC's prediction performance. The resulting gray-box MPC is compared against two white-box MPCs in a simulation environment. The results indicate an improvement in terms of maximum lateral tracking error from over \SI{2}{\mathbf{\meter}} to \SI{0.56}{\mathbf{\meter}}.

\end{abstract}


\section{Introduction}\label{kap:intro}
\begin{figure}[t]
   \centering
    \includegraphics[width=\linewidth]{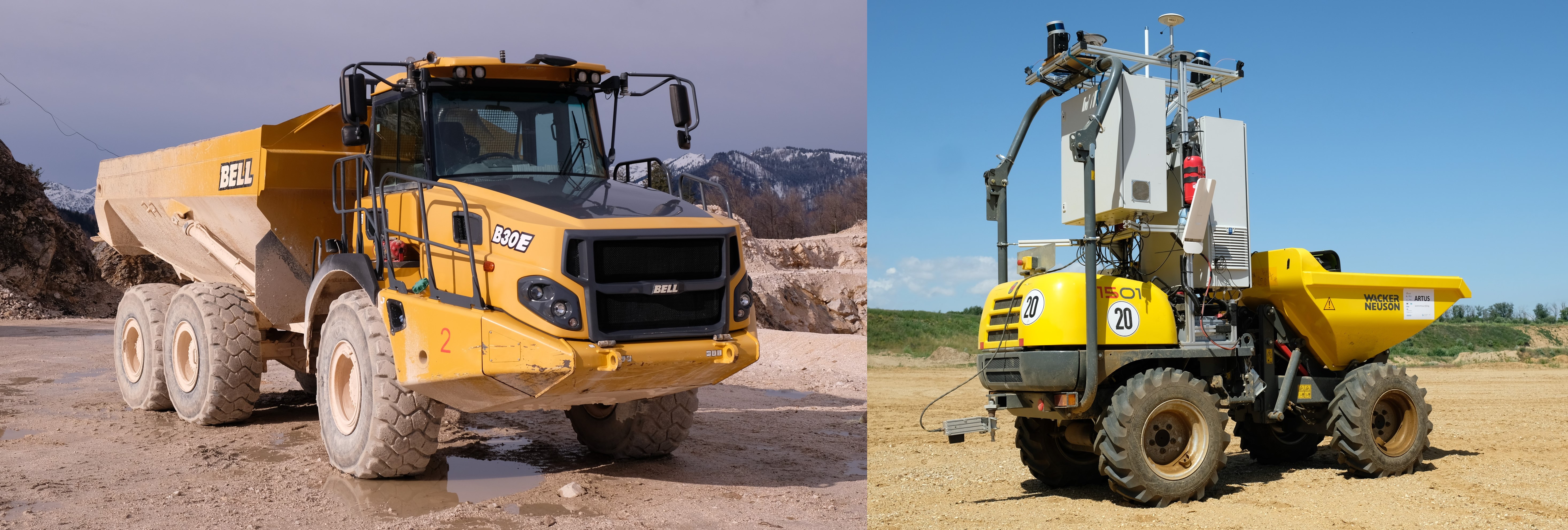}
	\caption{ADTs with equipment in a mining field. Image from \cite{shahirpour_design_2025}.}
\label{fig:Ng1} 
\end{figure}

Articulated dump trucks (ADTs) are among the most widely used vehicles in mining operations. These vehicles are designed and manufactured in different sizes for different use cases, such as the aforementioned mining operations and also in urban construction sites. Fig.~\ref{fig:Ng1} shows a full-sized and a compact ADT. 

The popularity of these vehicles, compared to front-steered (Ackermann) vehicles, stems from their ability to maintain maneuverability while transporting heavy loads, thanks to their steering mechanism. However, according to \cite{bellanca_why_2021}, these vehicles account for nearly half of the fatal mining accidents. Furthermore, the specific environment of mines makes operating these vehicles physically and mentally demanding~\cite{bauerle_mineworker_2018}, underscoring the relevance of automating them as a research topic. In the past decade, there have been many attempts to achieve autonomous ADTs. In the following, a short summary of these attempts is presented. 

In~\cite{altafini_path-tracking_1999},~\cite{dekker_experiments_2019}, and~\cite{ridley_load_2003}, feedback control methods are utilized for the autonomous operation of ADTs. Other non-predictive methods that have already been implemented include, but are not limited to, fuzzy control and the linear-quadratic regulator in works such as \cite{alshaer_modelling_2014} and~\cite{meng_lqr-ga_2019}. These methods are relatively easier to implement and deliver promising results for specific use cases in the mentioned studies. However, they are unable to predict future behavior, compensate for vehicle delays, or consider system constraints. This becomes specifically problematic at higher speeds, where the control delays in the vehicle actuation have more noticeable effects on vehicle behavior. As a result, these methods have limited applicability and performance. 

In works such as~\cite{shahirpour_systemidentifikation_2021} and~\cite{jeong_integrated_2023}, model predictive control (MPC) methods are presented that consider the vehicle delays. This is achieved by adding a delay model to the vehicle model used in the MPC. In \cite{shahirpour_design_2025}, the MPCs are tested in real experiments to evaluate their real-world performance. 

The mentioned works and the current studies in general mostly rely on kinematic vehicle models when it comes to model-based methods, although a dynamic model of the vehicle suitable for model-based algorithms does exist \cite{shahirpour_simulation_2022}. This is partly because kinematic models provide sufficient accuracy at lower speeds and in less dynamic driving scenarios, and they require far less effort to set up since they have considerably fewer parameters than dynamic models. As a result, dynamic models remain not widely used in the research despite the more accurate representation of vehicle motion. 

In a previous study \cite{shahirpour_simulation_2022}, we discussed how the sideslip angle plays an important role in accurately modeling the vehicle, and how neglecting it in kinematic models is one of the main reasons for their reduced accuracy at higher speeds and in curves, where the sideslip angle and its effect become noticeable. Consequently, we decided to include the sideslip angle in the vehicle's kinematic equations. However, this inclusion was only possible as a parameter rather than a dynamic variable, since a differentiable dynamic equation for the sideslip angle is not available within a purely kinematic framework.

Without a differentiable equation describing the sideslip angle, it was not possible to predict its behavior. As a result, within the MPC framework, the optimization only considered the current value of the sideslip parameter, without accounting for the factors influencing its future evolution or its impact on future vehicle motion. The modified MPC showed some improvement in \cite{shahirpour_sideslip_2023}, but more importantly indicated that a significant potential for further improvement remained. Consequently, in this paper, our objective is to establish a differentiable equation for the sideslip angle using learning-based methods.

In \cite{hewing_cautious_2020}, one of the first real-time MPC frameworks integrating learning-based models is presented, where Gaussian Process Regression (GPR) is used to account for model uncertainties and improve prediction accuracy of the vehicle dynamics in a race car. This work demonstrates how learning-based elements can be combined with physical models and utilized within an MPC framework to enhance prediction accuracy. 

Inspired by this idea, we implemented a GPR-based learning framework. In contrast to learning the residual dynamics of a full vehicle model, our proposed approach focuses on learning only the sideslip angle from the vehicle states: articulation angle, articulation angle rate, and longitudinal velocity $(\phi, \omega, v_x)$. The resulting learned model is then integrated into the kinematic vehicle model and used within the MPC framework. This new MPC is evaluated against previous MPC strategies in simulation. The objective of the proposed method is to reduce the gap between the kinematic and dynamic vehicle models.

In the following, the kinematic model of an ADT is presented in Section~\ref{sec:f-ADT_Model}. The learning process is then discussed in Section~\ref{sec:GPR}. Section~\ref{sec:MPC} presents a brief introduction to the MPC setup. In Section~\ref{sec:results-model}, the evaluation of the learned GPR function and the resulting gray-box kinematic model is discussed, while Section~\ref{sec:results-MPC} compares the new MPC with previously implemented ones. Finally, Section~\ref{sec:conclusion} presents the conclusion and outlook of this work. 

\section{Kinematic Model of an ADT}\label{sec:f-ADT_Model}
In this Section, we present both the kinematic model of an ADT and the kinematic-slip model, which is simply the kinematic model augmented with the sideslip angle as a parameter. In the next Section, the parameter $\alpha$ will be replaced with the GPR function to form the gray-box model. The kinematic bicycle model is a simple and frequently used model for model-based algorithms. In a bicycle model, the front and rear wheels are each modeled as a single wheel, located at the midpoint of their respective axles. 

\begin{figure}[t]
   \centering
    \includegraphics[width=\linewidth]{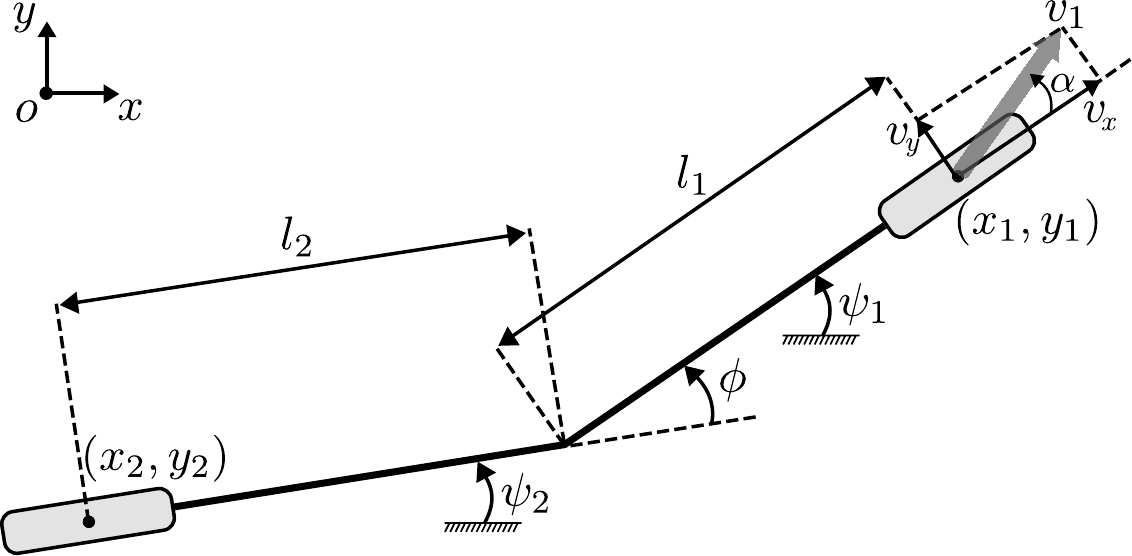}
	\caption{Kinematic configuration of an ADT with front sideslip based on \cite{shahirpour_simulation_2022}.}
\label{fig:Ng2} 
\end{figure}

The following model is based on \cite{delrobaei_design_nodate} and is extended to account for actuator delays. Fig.~\ref{fig:Ng2} shows the kinematic configuration of this vehicle. The state vector of this model is given as ${\mathbf{x}} = [{x}_1, {y}_1, \psi_1, \phi, \omega, v_x]^\top$. Here, $(x_1, y_1)$ is the position vector of the front axle in inertial coordinates, $\psi_1$ denotes the front yaw angle, while $\phi$ describes the articulation angle. The current speed is denoted by $v_x$ and the current articulation angle rate by $\omega$. The control input vector of this system is defined by $\mathbf{u} = [v_\ssd, \omega_\ssd]$, where $v_\ssd$ and $\omega_\ssd$ are the desired speed and articulation angle rate, respectively. The state equations of the ADT are given as:
\begin{subequations}\label{equ:kinmdl_forward_full-sized}
\begin{align}
\dot{x}_1 & = v_x \cos \psi_1,\ \  \dot{y}_1  = v_x \sin \psi_1, \tag{\theequation a,b}\label{equ:kinmdl_forward_full_a}\\ 
\dot{\psi}_1 & = \frac{\sin \phi}{l_2+l_1\cos\phi} v_x + \frac{l_2}{l_2+l_1\cos\phi} \omega, \label{equ:kinmdl_forward_full-sized_2C} \setcounter{equation}{2}\\
\dot{\phi} & = \omega, \\
\dot{\omega} & = (-\omega+k_{\omega}\omega_{\text{d}})/T_{\omega}, \\
\dot{v}_1 & = (-v_x+k_vv_\ssd)/T_v,
\end{align}
\end{subequations}
where ($k_{\omega}$, $k_v$) and ($T_{\omega}$, $T_v$) are the gains and time constants of steering and velocity, respectively. 

The kinematic model can be expanded to include the front sideslip angle as a parameter within the equations. The kinematic-slip model is presented in the following with the same state and input vector as the kinematic model~\cite{nayl_modeling_2013}:
\begin{subequations}\label{equ:kinslip}
\begin{align}
\dot{x}_1 & = v_x \cos (\psi_1+\alpha),\\  \dot{y}_1 & = v_x \sin (\psi_1+\alpha), \\ 
\dot{\psi}_1 & = \frac{\sin (\phi+\alpha)}{l_2+l_1\cos\phi} v_x + \frac{l_2}{l_2+l_1\cos\phi} \omega, \label{equ:kinmdl_forward_full-sized_2C} \setcounter{equation}{2}\\
\dot{\phi} & = \omega, \\
\dot{\omega} & = (-\omega+k_{\omega}\omega_{\text{d}})/T_{\omega}, \\
\dot{v}_x & = (-v_x+k_vv_\ssd)/T_v.
\end{align}
\end{subequations}

\section{Learning with Gaussian Process Regression}\label{sec:GPR}
\subsection{Basics of Gaussian Process Regression}
In this section, the basic idea behind learning the model mismatch with GPR is explained, inspired by \cite{hewing_cautious_2020}. As the focus of this work is not the GPR itself but its application, the introduction will be kept brief. 

Learning-based MPC approaches, such as those proposed by Hewing~\cite{hewing_cautious_2020}, address model inaccuracies by learning the full model mismatch using GPR and incorporating it as an additive correction term. These methods start by assuming that the model of the system can be presented as the following in a time-discrete format 
\begin{align}\label{equ:GPR_systemmodel}
{\mathbf{x}}(k+1) = f((x(k), u(k)) + g(x(k),u(k))+\omega(k),
\end{align}
where $x(k) \in \mathbb{R}^{n_x}$ is the system state and $u(k)\in \mathbb{R}^{n_u}$ is the system control input vector at the time step $k$. The function $f$ denotes the known and mathematically modeled part of the system dynamics, while $g$ is the unknown dynamics of the system. $\omega$ denotes zero-mean Gaussian process noise with variance $\sigma^2$, $\omega(k) \sim \mathcal{N}(0, \sigma_w^2)$. These dynamics will then be learned within the GPR framework to minimize model mismatch and improve accuracy.

\subsection{Interpretation of Model Mismatch via Sideslip Angle}
As discussed earlier, the main contribution of this work lies in learning a physically meaningful variable, namely the sideslip angle, instead of modeling the full model mismatch, i.e., residual dynamics. This is possible because the dominant effects of the model mismatch are largely attributable to the sideslip angle. This is presented in the following.  

When we look at the dynamic vehicle model for the position of the vehicle, we'll have the following equation, with $v$ denoting the magnitude of the velocity vector:
\begin{subequations}
\begin{align}
\dot{x}_1 & = v \cos (\psi_1+\alpha), \\
\dot{y}_1 & = v \sin (\psi_1+\alpha).
\end{align}
\end{subequations}
These equations can also be simply written as 
\begin{subequations}
\begin{align}
\dot{x}_1 & = 
\underbrace{v_x \cos\psi_1}_{\text{kinematic model}} 
\;-\;
\underbrace{v_y \sin\psi_1}_{\text{lateral velocity contribution}} ,\\
\dot{y}_1 & = 
\underbrace{v_x \sin\psi_1}_{\text{kinematic model}} 
\;+\;
\underbrace{v_y \cos\psi_1}_{\text{lateral velocity contribution}}.
\end{align}
\end{subequations}
Comparing these equations with the kinematic model for the position \eqref{equ:kinmdl_forward_full_a} illustrates the nature of the kinematic model, which is to assume that $v_y=0$ or, in other words, $\alpha=0$. As a result, if we were to apply Hewing's method here to learn the mismatch between the kinematic model and the dynamic model (or even the real vehicle), the mismatch would be primarily due to $v_y \sin\psi$ and $v_y \cos\psi$ for $\dot{x}$ and $\dot{y}$, respectively. 

Therefore, since the residual terms arising in the kinematic model can be interpreted as functions of the sideslip angle, we propose to learn $\alpha$ directly instead of learning $[u_y \sin\psi, u_y \cos\psi]^\top$. This reduces the complexity of the learning problem and improves data efficiency, as the regression task is constrained to a physically interpretable variable.

\subsection{Training the GPR model}\label{subsec:GPR_train}
To learn the sideslip angle, we don't use all the system states, but we limit the input to the most dominant ones in shaping the sideslip angle, which are: articulation angle $\phi$, articulation angle rate $\omega$, and speed $v_x$. Furthermore, this choice is motivated by practical considerations, as these variables are readily available from the vehicle’s CAN bus and navigation system, making the future real-world implementation easier while keeping the learning problem low-dimensional:
\begin{align}
    \alpha  = g_{\text{GP}}(\phi,\omega,v_x).
\end{align}
It is important to mention that this formulation represents a simplifying assumption in which the sideslip angle is modeled as an instantaneous function of the current states, neglecting its internal dynamics.

\begin{figure}[b!]
   \centering
   \includegraphics[
       width=\linewidth,
       trim=3cm 7cm 3cm 7cm,
       clip
   ]{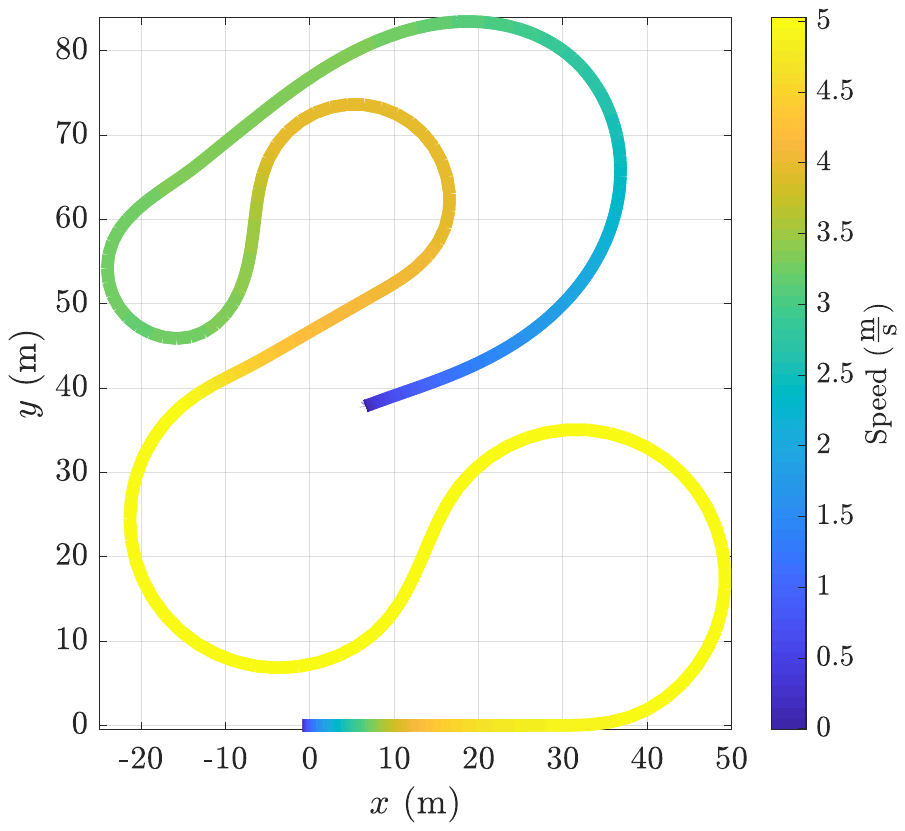}
   \caption{Reference trajectory from~\cite{shahirpour_simulation_2022}.}
   \label{fig:01_SollRef}
\end{figure}

As mentioned before, the main objective in this work is to improve the predictive accuracy of a kinematic model to approximate dynamic vehicle behavior. To this end, we collect vehicle data under representative driving conditions and use it to train a GPR model. The training data is generated in a simulation environment by driving the vehicle on a reference trajectory. The simulation environment uses a dynamic vehicle model and was validated in~\cite{shahirpour_simulation_2022}, and the reference trajectory was established in the same study from extensive operational vehicle data. This trajectory is designed to represent typical driving behavior in the target application domain. Fig.~\ref{fig:01_SollRef} illustrates this trajectory.

Since the focus of this work is not on Gaussian Process methodology itself but on its integration into the vehicle modeling and MPC framework, only the relevant implementation details are provided below to ensure reproducibility.

An exact GP formulation with an automatic relevance determination (ARD) squared exponential kernel is employed, allowing for individual length scales for each input dimension. The hyperparameters, namely the signal variance~$\sigma_F^2$, the noise variance~$\sigma$, and the input length scales~$l = [l_\phi, l_\omega,l_{v_x}]$,  are selected manually to ensure stable and smooth model behavior, with~$\sigma_F = 1$,~$\sigma= 0.001$, and~$l_\phi = l_\omega = l_{v_x} = 6$. Training is performed using standard GP regression with Cholesky decomposition~\cite{rasmussen_gaussian_2010}.

When the learned sideslip angle function $g$ is incorporated into the system model, the new gray-box model becomes 
\begin{subequations}\label{equ:GPRModel}
\begin{align}
\dot{x}_1 & = v_1 \cos (\psi_1+g_{\text{GP}}(\phi, \omega,v_x)),\\  \dot{y}_1 & = v_1 \sin (\psi_1+g_{\text{GP}}(\phi, \omega,v_x)), \\ 
\dot{\psi}_1 & = \frac{\sin (\phi+g_{\text{GP}}(\phi, \omega,v_x))}{l_2+l_1\cos\phi} v_1 + \frac{l_2}{l_2+l_1\cos\phi} \omega, \label{equ:kinmdl_forward_full-sized_2C} \setcounter{equation}{2}\\
\dot{\phi} & = \omega, \\
\dot{\omega} & = (-\omega+k_{\omega}\omega_{\text{d}})/T_{\omega}, \\
\dot{v}_x & = (-v_1+k_vv_\ssd)/T_v.
\end{align}
\end{subequations}
\section{Trajectory-Following MPC}\label{sec:MPC}
The objective of the MPC is to find the optimal control trajectory that minimizes the distance between the vehicle's front axle and a given reference trajectory $\mathbf{x}^\star$.  

The MPC framework requires linear and time-discrete ($\dashedph_\sstd$) system models. This means to establish the linear and time-discrete state and input matrices $\mathbf{A}_\sstd, \mathbf{B}_\sstd$ for the system. To this goal, the Jacobian matrix can be employed on system equations (denoted as $\mathbf{f}$) to achieve the linear system matrices $\mathbf{A}, \mathbf{B}$. This will be done each time the MPC is called at each operating point (OP) as follows
\begin{align}
    &\mathbf{A}=  \frac{\partial\mathbf{f}}{\partial\mathbf{x}}\Bigr|_{\substack{\mathbf{x}_{\text{op}},\mathbf{u}_{\text{op}}}},\ 
    \mathbf{B} =  \frac{\partial\mathbf{f}}{\partial \mathbf{u}}\Bigr|_{\substack{\mathbf{x}_{\text{op}},\mathbf{u}_{\text{op}}}}.
\end{align}
After linearization and discretization of each model with sampling time $T_{\text{s}}$ using zero-order hold, the linear and time-discrete system model is established
\begin{align}\label{equ:lin_system}
\mathbf{x}_\sstd(k+1)=
\mathbf{A}_\sstd(k)\mathbf{x}_\sstd(k) + \mathbf{B}_\sstd(k)\mathbf{u}_\sstd(k).
\end{align}
In the following, the cost function of the optimization problem is presented
\begin{subequations}    \label{eq:Cost_function_mpc}
    \begin{alignat}{2}
        & \mathmakebox[\widthof{$\min_{x}$}][c]{\min_{\mathbf{u}_\sstd(\cdot|k)}}            &   \qquad     & \sum^{N_{\text{c}}}_{i=1} \| \mathbf{x}_\sstd(k+i|k)-\mathbf{x}^\star(k+i|k) \|^2_{\mathbf{Q}} \, + \\
        &&&\, + \|\mathbf{u}_\sstd(k+i-1|k)-\mathbf{u}_\sstd(k+i-2|k)\|^2_{\mathbf{R}}\notag  \\
        & \mathmakebox[\widthof{$\min_{x}$}][c]{\text{s.t.}} &&  \mathbf{x}_\sstd(\cdot|k) \in \mathbb{X},\quad \mathbf{u}_\sstd(\cdot|k) \in \mathbb{U},
    \end{alignat}
\end{subequations}
where ${\mathbf{Q}}$ and ${\mathbf{R}}$ denote the weight matrices, and $\mathbb{X}$ and $\mathbb{U}$ denote the set of feasible states and control inputs, respectively. 

\section{Simulation Results: Evaluation of the GPR Model}\label{sec:results-model}
Before integrating the proposed model into the MPC framework, we first evaluate its standalone performance. This is done by evaluating its generalization performance on previously unseen and highly dynamic test data.

\begin{figure}[b]
   \centering
    \includegraphics[width=\linewidth]{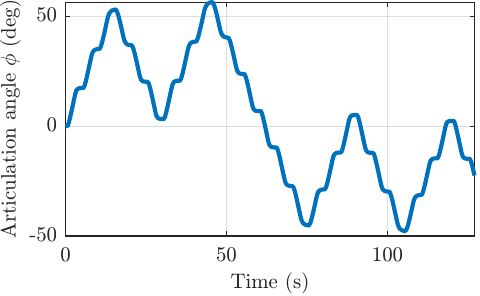}
	\caption{Articulation angle $\phi$ profile for creating the test track.} 
	\label{fig:03_phi_profile}
\end{figure}

For this goal, the ADT is driven in a simulation on a random track at speeds between \SI{4}{} and \SI{5}{\meter\per\second}, with an articulation angle profile shown in Fig.~\ref{fig:03_phi_profile}, to test the GPR model under extreme conditions. As Fig.~\ref{fig:03_phi_profile} indicates, the reference articulation angle in the test track reaches above \SI{50}{\degree} in some instances, which is more than the limit of \SI{43}{\degree} in the test data, contributing to more dynamic and extreme driving for testing. 

\begin{figure}[t]
   \centering
    \includegraphics[width=\linewidth]{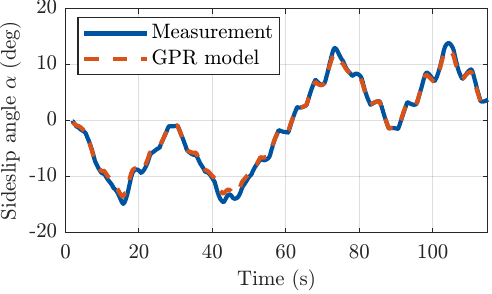}
	\caption{Measurement (dynamic model) vs. GPR model prediction for sideslip angle $\alpha$ on the test data.} 
	\label{fig:04_sideslip_plot}
\end{figure}

Fig.~\ref{fig:04_sideslip_plot} compares the measured (from the dynamic model) and predicted sideslip angle~$\alpha$ as the vehicle drives on the test track. As the Figure demonstrates, the model provides a close prediction, especially when $\alpha < \SI{10}{\degree}$. This is primarily due to the fact, that the training data also only included data points with a maximum sideslip angle of \SI{7.5}{\degree}. In other words, the training data was not sufficiently dynamic compared to the test data. 

\begin{figure}[t]
   \centering
    \includegraphics[width=\linewidth]{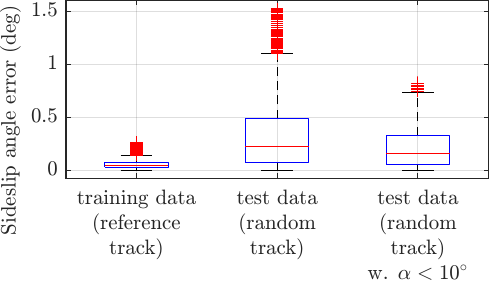}
	\caption{Box plot of the absolute sideslip angle $\alpha$ prediction error by the GPR model.} 
	\label{fig:05_boxplot_comp_sideslip}
\end{figure}

Fig.~\ref{fig:05_boxplot_comp_sideslip} shows the box plot results for the absolute error in sideslip angle for evaluation both with test and train data. Furthermore, it shows the evaluation on test data with sideslip angles less than \SI{10}{\degree}. As the Figure shows, the model has its best performance, as expected, when predicting the sideslip angle on the training data. On the test data, the results are nevertheless promising, with the average absolute error under \SI{0.5}{\degree}. When the presentation is limited to sideslip angles below \SI{10}{\degree}, which is closer to the range of the training data, the results improve.

This concludes the training of the GPR model $g_{\text{GP}}(\phi,\omega,v_x))$. We have now successfully shown that the GPR model has promising potential in learning the sideslip angle using the state set $(\phi,\omega,v_x)$. 

In the next step, to evaluate to what extent we have achieved our initial goal, which was to minimize the gap between the dynamic and the kinematic model by introducing learning to the kinematic model, we now compare the gray-box model \eqref{equ:GPRModel} with the dynamic model before we move on to the MPCs. 

For this comparison, we do not compare the position vectors of the kinematic and dynamic models, as they are the results of integration and any errors between the underlying terms accumulate over time, not leading to a meaningful comparison. Instead, we compare the velocity vector between the two models, in particular the lateral velocity of the vehicle in the front axle ${v}_y$. The lateral velocity is readily available from the dynamic model as a state. The kinematic model does not directly provide this variable, but the following equation yields the lateral velocity
\begin{subequations}
\begin{align}
\dot{v}_{y_\text{kin}} & = v \sin \alpha,\\
& = v \sin(g_{\text{GP}}(\phi, \omega,v_x)).
\end{align}
\end{subequations}
To preserve the model's kinematic structure, the longitudinal velocity $v_x$ is used instead of the velocity magnitude $v$ in this comparison. This ensures that the model remains consistent within a kinematic formulation augmented by a learned component and avoids introducing dependencies on inherently dynamic variables. 

\begin{figure}[b!]
   \centering
    \includegraphics[width=\linewidth]{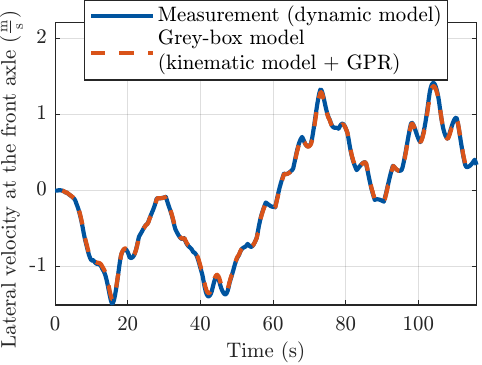}
	\caption{Lateral velocity at the front axle $v_y$ comparison between the dynamic model and the gray-box model on the test data.} 
	\label{fig:06}
\end{figure}
\begin{figure}[b!]
   \centering
    \includegraphics[width=\linewidth]{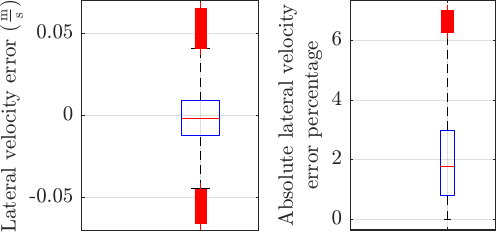}
	\caption{Lateral velocity error and absolute lateral velocity error percentage comparison between the dynamic model and the gray-box model on the test data.} 
	\label{fig:07}
\end{figure}

Fig.~\ref{fig:06} shows the comparison between the two lateral velocities, while Fig.~\ref{fig:07} illustrates the error between the two. While the error value is meaningful, the error percentage provides a better understanding of the performance. Therefore, we have also included the error percentage after removing data points of the lateral velocities smaller than $\epsilon<\SI{0.3}{\meter\per\second}$ to avoid division by small values.

As the Figures indicate, the gray-box model is able to predict the lateral velocity by utilizing the GPR model of the sideslip angle with an average error under $2 \%$ and a maximum under $7 \%$. It's important to mention that the gray-box model achieves these results on the test track, which, as mentioned earlier, represents a more dynamic track than the training data. This is also evident here, as the most noticeable errors between the two values in Fig.~\ref{fig:06} occur at lateral velocities that correspond to sideslip angles over \SI{10}{\degree}, which was not included in the training data. 

It is also important to emphasize that white-box kinematic models assume $v_y =\SI{0}{\meter\per\second}$ by assuming $\alpha = \SI{0}{\degree}$. In other words, the established gray-box model has reduced the gap between the kinematic models and dynamic models. In the next Section, we investigate to what extent this improvement actually contributes to the MPC performance. 

\begin{table}[!b]
	\begin{center}
		\caption{MPC parameters}		\label{tb:mpc-params-Bell}
		\begin{tabular}{l l r}
		\hline
			\textbf{Parameter} & \textbf{Symbol} & \textbf{Value}   \\\hline
			Vehicle lengths & $l_1,l_2$ & $1.36,\,$\SI{3.65}{\meter}  \\
			Speed constants & $T_v, k_v$ & \SI{1.25}{\second}, $1$ \\ 
			Steering constants  & $T_\omega, k_\omega$ & \SI{0.5}{\second}, $1$ \\
		    Weight matrix  & $\mathbf{Q}$ & $\text{diag}([100,100,0,0,0,0])$ \\
		    Weight matrix  & $\mathbf{R}$ & $\text{diag}([1,1])$ \\
		    Further parameters  & $T_{\text{d}}, T_{\text{s}}, N_{\text{c}}$ & \SI{0.5}{\second}, \SI{0.3}{\second}, $20$ \\
                  MPC's state constraints  & $|\phi|_\smax$ & \SI{42}{\deg}\\
            MPC's input constraints  & $|\omega_\ssd|_\smax$,$|v_\ssd|_\smax$ & \SI{12}{\deg\per\second}, \SI{8}{\meter\per\second} \\
            \hline
		\end{tabular}
	\end{center}
\end{table}

\section{Simulation Results: Evaluation of the MPCs}\label{sec:results-MPC}
The goal of this simulation experiment is to compare the performance of the three presented ADT models within an MPC framework. The first vehicle model is presented in \eqref{equ:kinmdl_forward_full-sized} and is called the white-box kinematic model. The second model is presented in \eqref{equ:kinslip} and is called the white-box kinematic-slip model, and the third model is presented in \eqref{equ:GPRModel} and is called the gray-box kinematic model. These models are used in three MPCs, each named after the model. These MPCs have identical properties listed under table~\ref{tb:mpc-params-Bell}. The performance of these MPCs is then compared in terms of the lateral error to the reference trajectory that was presented in Section~\ref{subsec:GPR_train} and illustrated in Fig.~\ref{fig:01_SollRef}. 

\begin{figure}[t!]
   \centering
    \includegraphics[width=\linewidth]{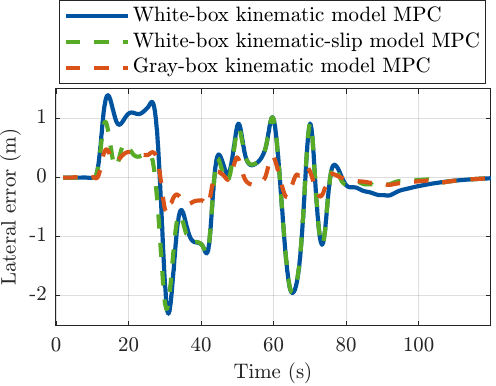}
	\caption{Lateral error between the reference trajectory and the front axle for the three MPCs.} 
	\label{fig:08}
\end{figure}

\begin{figure}[t!]
   \centering
    \includegraphics[width=\linewidth]{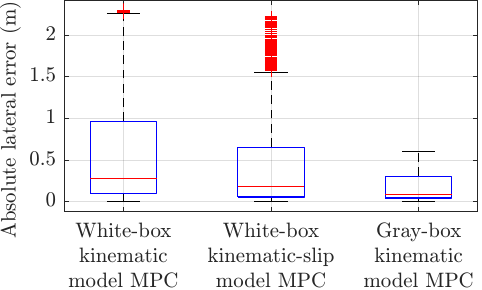}
	\caption{Absolute lateral error between the reference trajectory and the front axle for the three MPCs.} 
	\label{fig:09}
\end{figure}

Fig.~\ref{fig:08} shows the lateral error between the vehicle's front axle and the reference trajectory when driving with each MPC, and Fig.~\ref{fig:09} illustrates the box plot of the absolute lateral error. As Fig.~\ref{fig:08} indicates, the kinematic-slip MPC already improves the results compared to the white-box kinematic model; however, the improvement is expectedly limited since the inclusion of the sideslip angle is purely as a parameter. In practical terms, it means that the model does not know what contributes to the value of the sideslip angle and cannot predict it. The gray-box model, on the other hand, has the GPR function for the sideslip angle and utilizes that function to better consider current and future sideslip angles in the vehicle movement. We can also see that the maximum error has dropped to \SI{0.56}{\meter} for the MPC with the gray-box model, while the maximum error is over \SI{2}{\meter} for the other MPCs whereas the other MPCs have maximum errors over \SI{2}{\meter}. Fig.~\ref{fig:09} also shows how using the gray-box model MPC has improved the results not only in terms of the maximum error but also in error distribution.

These results demonstrate that using a gray-box model with a GPR function for the sideslip angle within an MPC framework improves the results by reducing the gap between the kinematic and the dynamic model without introducing extra parametrization difficulties and while keeping the kinematic nature of the model and MPC.

While the results indicate promising potential, it is important to mention that this method is still in its theoretical and simulation phase. For instance, while the new gray-box model reduces the gap between the kinematic and dynamic models and removes parametrization difficulties, it still requires training data. Addressing these issues will be the topic of our future work.


\section{Conclusion}\label{sec:conclusion}
In this work, we proposed a Gaussian Process Regression (GPR)-based method to reduce the gap between kinematic and dynamic ADT models. We show that this mismatch can be primarily attributed to the sideslip angle, which is therefore selected as the learned variable and incorporated into the kinematic model to form a gray-box model. 

The proposed gray-box model reduces the gap between kinematic and dynamic behavior, as demonstrated by improved lateral velocity prediction, a quantity that is assumed to be zero in the kinematic formulation. The results show an average lateral velocity prediction error below $2\%$ for the gray-box model. Furthermore, when embedded in an MPC framework for trajectory tracking, the proposed model yields a noticeable reduction in lateral tracking error.

Learning a physically meaningful variable using GPR provides an alternative approach to residual-based learning, enabling a more structured and interpretable modeling approach. In addition, the proposed gray-box formulation improves predictive accuracy in the MPC without requiring a full dynamic model, therefore avoiding the associated parametrization and modeling complexity while preserving compatibility with existing MPC frameworks.

Future work will focus on extending the method to experimental data and testing, and investigating adaptive learning strategies for varying operating conditions and particularities. 


\bibliographystyle{IEEEtran}
\bibliography{IEEEabrv, references}
\end{document}